\PassOptionsToPackage{table}{xcolor} 
\documentclass[11pt,a4paper]{article}
\usepackage[hyperref]{acl2020}
\usepackage{enumerate}
\usepackage{hyperref}
\usepackage{url}
\usepackage{tikz-dependency}
\usepackage[title]{appendix}
\usepackage{afterpage}
\usepackage{nicefrac}
\usepackage{algorithm,amsmath,algpseudocode}
\usepackage{subcaption}
\usepackage{multirow}
\usepackage{array,booktabs}
\usepackage{xcolor}
\usepackage{times}
\usepackage{enumerate}
\usepackage{latexsym}
\usepackage{url}
\usepackage{amsmath}
\usepackage{graphicx}
\usepackage{float}
\usepackage{placeins}
\usepackage[printwatermark]{xwatermark}
\aclfinalcopy 
\def\aclpaperid{5} 

\usepackage{algorithm,amsmath,algpseudocode}
\usepackage{subcaption}
\usepackage{multirow}
\usepackage{siunitx}
\usepackage{array,booktabs}
\usepackage[title]{appendix}
\usepackage{afterpage}
\usepackage{nicefrac}

\title{Distilling Neural Networks for Greener and Faster Dependency Parsing}

\author{Mark Anderson\qquad Carlos G\'{o}mez-Rodr\'{i}guez\\
  Universidade da Coru\~na, CITIC \\
  FASTPARSE Lab, LyS Research Group, \\ 
  Departamento de Ciencias de la Computaci\'{o}n y Tecnolog\'{i}as de la Informaci\'{o}n \\
  Campus Elvi\~{n}a, s/n, 15071 
  A Coru\~{n}a, Spain\\
  {\texttt \{m.anderson,carlos.gomez\}@udc.es}}

\date{}
\newcommand{\carlos}[1]{\textcolor{black}{#1}}

\newcommand{\john}[1]{\textcolor{black}{#1}}
\definecolor{LightMagenta}{RGB}{249, 249, 255}
\definecolor{deppink}{HTML}{8d3b72}

\begin{document}

\maketitle
\begin{abstract}
The carbon footprint of natural language processing research has been increasing in recent years due to its reliance on large and inefficient neural network implementations. Distillation is a network compression technique which attempts to impart knowledge from a large model to a smaller one. We use \textit{teacher-student} distillation to improve the efficiency of the Biaffine dependency parser which obtains state-of-the-art performance with respect to accuracy and parsing speed \citep{dozat20161}. When distilling to 20\% of the original model's trainable parameters, we only observe an average decrease of $\sim$1 point for both UAS and LAS across a number of diverse Universal Dependency treebanks while being 2.30x (1.19x) faster 
 than the baseline model 
 on CPU (GPU) 
 at inference time. We also observe a small increase in performance when compressing to 80\% for some treebanks. Finally, through distillation we attain a parser which is not only faster but also more accurate than the fastest modern parser on the Penn Treebank.
\end{abstract}

\section{Introduction}
Ethical NLP research has recently gained attention 
\citep{kurita2019measuring,sun2019mitigating}. 
For example, the environmental cost of AI research has become a focus of the community, especially with regards to the development of deep neural networks \citep{schwartz2019green,strubell2019energy}. Beyond developing systems to be greener, increasing the efficiency of models makes them more cost-effective, which is a compelling argument even for people who might downplay the extent of anthropogenic climate change.

In conjunction with this push for greener AI, NLP practitioners have turned to the problem of developing models that are not only accurate but also efficient, so as to make them more readily deployable across different machines with varying computational capabilities \citep{strzyz2019viable,clark2019bam,vilares2019better,junczys2018marian}. This is in contrast with the 
recently popular principle of \textit{make it bigger, make it better} \citep{devlin2018bert,radford2019language}. 

Here we explore \textit{teacher-student} distillation as a means of increasing the efficiency of neural network systems used to undertake a core task in NLP, dependency parsing. To do so, we take a state-of-the-art Biaffine parser from \citet{dozat20161}. The Biaffine parser is not only one of the most accurate parsers, it is the fastest implementation by almost an order of magnitude among state-of-the-art performing parsers.

\paragraph{Contribution} We utilise \textit{teacher-student} distillation to compress Biaffine parsers trained on a 
diverse subset of 
Universal Dependency (UD) treebanks. We find that distillation maintains accuracy performance close to that of the full model and obtains far better accuracy than simply implementing equivalent model size reductions by changing the parser's network size and training normally. 
Furthermore, we can compress a parser to 20\% of its trainable parameters with minimal loss in accuracy and with a speed 2.30x (1.19x) faster than that of the original model on CPU (GPU).

\section{Dependency parsing}
\begin{figure}[htpb!]
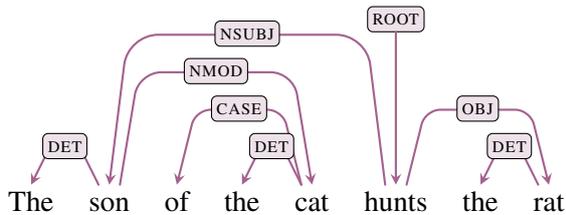

\centering
\begin{dependency}[edge style={deppink!80, thick},label style={fill=deppink!15},edge slant=7]
\begin{deptext}[column sep=0.75em,ampersand replacement=\^]
The \^ son \^ of \^ the \^ cat \^ hunts \^ the \^ rat \\
\end{deptext}
\depedge{2}{1}{\textsc{det}}
\depedge{6}{2}{\textsc{nsubj}}
\depedge{5}{3}{\textsc{case}}
\depedge{5}{4}{\textsc{det}}
\depedge{2}{5}{\textsc{nmod}}
\depedge{8}{7}{\textsc{det}}
\depedge{6}{8}{\textsc{obj}}
\deproot[edge unit distance=3.5ex]{6}{\textsc{root}}
\end{dependency}
\caption{Dependency tree example.}\label{fig:dep}
\end{figure}

Dependency parsing is a core NLP task where the syntactic relations of words in a sentence are encoded as a well-formed tree with each word attached to a head via a labelled arc. Figure \ref{fig:dep} shows an example of such a tree. The syntactic information attained from parsers has been shown to benefit a number of other NLP tasks such as 
relation extraction \citep{zhang2018graph}, machine translation \citep{chen2018syntax},
and sentiment analysis \citep{poria2014sentic,vilares2017universal}.


\subsection{Current parser performance}

Table \ref{tab:current-speeds} shows performance details of current state-of-the-art dependency parsers on the English Penn Treebank (PTB) with predicted POS tags from the Stanford POS tagger \citep{marcus1993building,toutanova2003feature}. The Biaffine parser of \citet{dozat20161} offers the best trade-off between accuracy and parsing speed with the HPSG parser of \citet{zhou2019head} achieving the absolute best reported accuracy but with a reported parsing speed of roughly one third of the Biaffine's parsing speed. It is important to note that direct comparisons between systems with respect to parsing speed are wrought with compounding variables, e.g. different GPUs \carlos{or CPUs} used, \carlos{different number of CPU cores,} different batch sizes, and often hardware is not even reported. 

\begin{table*}[htpb!]
\small
    \centering
    \tabcolsep=.15cm  
   
    \begin{tabular}{l|cc|cc}
     \multicolumn{1}{c}{} & \multicolumn{2}{c}{\textbf{speed (sent/s)}} \\
       \multicolumn{1}{c}{}  & \multicolumn{1}{c}{\textbf{GPU}} & \multicolumn{1}{c}{\textbf{CPU}} & \textbf{UAS} & \textbf{LAS} \\\hline&&&&\\[-2ex]
    
    Pointer-TD \citep{ma2018stack}& - & 10.2$^{\dagger}$ & 95.87$^{\dagger}$ & 94.19$^{\dagger}$ \\
    Pointer-LR \citep{fernandez2019left}& -  & 23.1$^{\dagger}$ & 96.04$^{\dagger}$ & 94.43$^{\dagger}$ \\
    HPSG \citep{zhou2019head} & 158.7$^{\dagger}$ & - & 96.09$^{\dagger}$ & 94.68$^{\dagger}$ \\
    BIST - Transition \citep{kiperwasser2016simple} & - & 76$\pm$1$^{\ddagger}$ & 93.9$^{\dagger}$ & 91.9$^{\dagger}$\\
    BIST - Graph \citep{kiperwasser2016simple}& - & 80$\pm$0$^{\ddagger}$ & 93.1$^{\dagger}$ & 91.0$^{\dagger}$ \\
    Biaffine \citep{dozat20161} & 411$^{\dagger}$ & - &95.74$^{\dagger}$ &  94.08$^{\dagger}$\\ 
    CM \citep{chen2014fast1}& - & 654$^{\dagger}$ & 91.80$^{\dagger}$ & 89.60$^{\dagger}$  \\
    SeqLab \citep{strzyz2019viable} & 648$\pm$20$^{\ddagger}$ & 101$\pm$2$^{\ddagger}$ & 93.67$^{\ddagger}$ & 91.72$^{\ddagger}$   
    \\\hline\hline\rowcolor{LightMagenta}&&&&\\[-2ex]
    \rowcolor{LightMagenta}
    UUParser \citep{smith201882} &-& 42$\pm$1& 94.63 & 92.77 \\
    \rowcolor{LightMagenta}
    Biaffine (PyTorch) & 1003$\pm$3 & 53$\pm$0 & 95.74 & 94.07 \\
    \rowcolor{LightMagenta}
    SeqLab  & 1064$\pm$13 & 99$\pm$1 & 93.46 & 91.49  \\
    \rowcolor{LightMagenta}
    \hline&&&&\\[-2ex]
    \rowcolor{LightMagenta}
    Biaffine-D20 & 1189$\pm$4&  391$\pm$2 & 92.84 & 90.73   \\
    \rowcolor{LightMagenta}
    \textbf{Biaffine-D40} & \textbf{1153$\pm$3} &  \textbf{96$\pm$0} & \textbf{94.59} & \textbf{92.64}  \\
    \rowcolor{LightMagenta}
    Biaffine-D60 & 1112$\pm$6&  71$\pm$1 & 94.78 & 92.86  \\
    \rowcolor{LightMagenta}
    Biaffine-D80 & 1033$\pm$5&  61$\pm$0 & 94.84 & 92.95  \\ 
    \end{tabular}
    \caption{Speed and accuracy performance for state-of-the-art parsers and parsers from our distillation method, Biaffine-D$\pi$ compressing to $\pi$\% of the original model, for the English PTB with POS tags predicted from the Stanford POS tagger. 
    \carlos{In the first table block,} $\dagger$ denotes values taken from the original paper and $\ddagger$ from \citet{strzyz2019viable}. Values with no superscript \carlos{(corresponding to the models in the shaded area, i.e. the second and third blocks)} are from running the models on our system locally with a single CPU core for both CPU and GPU speeds (averaged over 5 runs) and with a batch size of 256 (excluding UUParser which doesn't support batching) with GloVe 100 dimension embeddings. 
    }
    \label{tab:current-speeds}
\end{table*}

We therefore run a subset of parsers locally to achieve speed measurements in a controlled environment, also shown in Table \ref{tab:current-speeds}: we compare a PyTorch implentation of the Biaffine parser (which runs more than twice as fast as the reported speed of the original implementation); the UUParser from \citet{smith201882} which is one of the leading parsers for Universal Dependency (UD) parsing; a sequence-labelling dependency parser from \citet{strzyz2019viable} which has the fastest reported parsing speed amongst modern parsers; and also distilled Biaffine parsers from our implementation described below. All speeds measured here are with the system run with a single 
\carlos{CPU core}
for both GPU and CPU 
\carlos{runs.}\footnote{\carlos{This is for ease of comparability. Parsing can trivially be parallelised by allocating sentences to different cores, so speed per core is an informative metric to compare parsers \citep{hall-etal-2014-sparser}.}}

\paragraph{Biaffine parser} is a graph-based parser extended from the graph-based BIST parser \citep{kiperwasser2016simple} to use a biaffine attention mechanism which pairs the prediction of edges with the prediction of labels. This results in a fast and accurate parser, as described above, and is used as the parser architecture for our experiments. More details of the system can be found in \citet{dozat20161}.
\section{Network compression}
Model compression has been under consideration for almost as long as neural networks have been utilised, e.g. \citet{lecun1990} introduced a pruning technique which removed weights based on a locally predicted contribution from each weight so as to minimise the perturbation to the error function. More recently, \citet{han2015learning} introduced a means of pruning a network up to 40 times smaller with minimal affect on performance. \citet{hagiwara1994} and \citet{wan2009} utilised magnitude-based pruning to increase network generalisation. More specific to NLP, \citet{see2016} used absolute-magnitude pruning to compress neural machine translation systems by 40\% with minimal loss in performance. However, pruning networks leaves them in an irregularly sparse state which cannot be trivially re-cast into less sparse architectures. Sparse tensors could be used for network layers to obtain real-life decreases in computational complexity, however, current deep learning libraries lack this feature. \citet{anwar2017} introduced structured pruning to account for this, but this kernel-based technique is restricted to convolutional networks. More recently \citet{voita2019analyzing} pruned the heads of the attention mechanism in their neural machine translation system and found that the remaining heads were linguistically salient with respect to syntax, suggesting that pruning could also be used to undertake more interesting analyses beyond merely compressing models and helping generalisation.

 \citet{ba2014} and \citet{hinton2015} developed distillation as a means of network compression from the work of \citet{bucilua2006model}, who compressed a large ensemble of networks into one smaller network. Similar and more recent work, used this method of compressing many models into one to achieve state-of-the-art parsing performance \cite{kuncoro-etal-2016-distilling}.   \textit{Teacher-student} distillation is the process of taking a large network, the \textit{teacher}, and transferring its knowledge to a smaller network, the \textit{student}. \textit{Teacher-student} distillation has successfully been exploited in NLP for machine translation, language modelling, and speech recognition \citep{kim2016sequence,yu2018device,lu2017knowledge}. Beyond that it has also been successfully used in conjunction with exploring structured linguistic prediction spaces \cite{liu-etal-2018-distilling}. Latterly, it has also been used to distill task-specific knowledge from BERT \citep{tang2019distilling}. 

Other compression techniques have been used such as low-rank approximation decomposition \citep{yu2017compressing}, vector quantisation \citep{wu2016quantized}, and Huffman coding \citep{han2015deep}. For a more thorough survey of current neural network compression methods see \citet{cheng2017survey}. 

\section{Teacher-student distillation}
The essence of model distillation is to train a model and subsequently use the patterns it learnt to influence the training of a smaller model. For \textit{teacher-student} distillation, the smaller model, the \textit{student}, explicitly uses the information learnt by the larger original model, the \textit{teacher}, by comparing the distribution of each model's output layer.  We use the Kullback-Leibler divergence to calculate the loss between the teacher and the student:
\begin{equation}
    \mathcal{L}_{KL}= -\sum_{t \in b}\sum_{i} P(\textbf{x}_i)\log\frac{P(\textbf{x}_i)}{Q(\textbf{x}_i)}
\end{equation}
where $P$ is the probability distribution from the teacher's softmax layer, $Q$ is the probability distribution from the student's, and $\textbf{x}_i$ is input vector to the softmax corresponding to token $w_i$ of a given tree $t$ for all trees in batch $b$. 

For our implementation, there are two probability distributions for each model, one for the arc prediction and one for the label prediction.  By using the distributions of the teacher rather than just using the predicted arc and label, the student can learn more comprehensively about which arcs and labels are very unlikely in a given context, i.e. if the teacher makes a mistake in its prediction, the distribution might still carry useful information such as having a similar probability for $y_g$ and $y_p$ which can help guide the student better rather than just learning to copy the teacher's predictions.

In addition to the loss with respect to the teacher's distributions, the student model is also trained using the loss on the gold labels in the training data.  We use categorical cross entropy to calculate the loss on the student's predicted head classifications:
\begin{equation}
    \mathcal{L}_{CE}= -\sum_{t \in b}\sum_{i} \log p(h_i|\textbf{x}_i)
\end{equation}
where $h_i$ is the true head position for token $w_{i}$, corresponding to the softmax layer input vector $\textbf{x}_i$, of tree $t$ in batch $b$. 
Similarly, categorical cross entropy is used to calculate the loss on the predicted arc labels for the student model. The total loss for the student model is therefore:
\begin{align}
    \mathcal{L} = \mathcal{L}_{KL}&(T_{h}, S_{h}) + \mathcal{L}_{KL}(T_{lab}, S_{lab})\nonumber\\ &+ \mathcal{L}_{CE}(h) + \mathcal{L}_{CE}(lab)
\end{align}
where $\mathcal{L}_{CE}(h)$ is the loss for the student's predicted head positions, $\mathcal{L}_{CE}(lab)$ is the loss for the student's predicted arc label, $\mathcal{L}_{KL}(T_h, S_h)$ is the loss between the teacher's probability distribution for arc predictions and that of the student, and $\mathcal{L}_{KL}(T_{lab}, S_{lab})$ is the loss between label distributions. This combination of losses broadly follows the methods used in \citet{tang2019distilling} but is altered to fit the Biaffine parser.


\section{Methodology}
We train Biaffine parsers and apply the \textit{teacher-student} distillation method to compress these models into a number of different sizes for a number of Universal Treebanks v2.4 (UD) \citep{nivre2019}. We use the hyperparameters from \citet{dozat20161}, but use a PyTorch implementation for our experiments which obtains the same parsing results and runs faster than the reported speed of the original (see Table \ref{tab:current-speeds}).\footnote{
The implementation can be found at \texttt{github.com/zysite/biaffine-parser}. Beyond adding our distillation method, we also included the Chu-Liu/Edmonds' algorithm, as used in the original, to enforce well-formed trees.} The hyperparameter values can be seen in Table \ref{tab:experimental_hyperparameters} in the Appendix. During distillation dropout is not used as in earlier experiments with dropout performance was hampered. And subsequent work on distillation which uses dropout also didn't perform well, but it isn't clear if this is the cause of the poorer performance, e.g. different treebanks were used, UPOS tags weren't, and no pre-trained embeddings were used \cite{dehouck2020}. Beyond lexical features, the model only utilises universal part-of-speech (UPOS) tags. Gold UPOS tags were used for training and at runtime. Also, we used gold sentence segmentation and tokenisation. We opted to use these settings to compare models under homogeneous settings, so as to make reproducibility of and comparability with our results easier.

\paragraph{Data} We use the subset of UD treebanks suggested by \citet{de2017old} from v2.4, so as to cover a wide range of linguistic features, linguistic typologies, and different dataset sizes. We make some changes as this set of treebanks was chosen from a previous UD version. We exchange Kazakh with Uyghur because the Kazakh data does not include a development set and Uyghur is a closely related language. We also exchange Ancient-Greek-Proiel for Ancient-Greek-Perseus because it contains more non-projective arcs (the number of arcs which cross another arc in a given tree) as this was the original justification for including Ancient Greek. Further, we follow \citet{smith2018investigation} and exchange Czech-PDT with Russian-GSD.  We also included Wolof as African languages were wholly unrepresented in the original collection of suggested treebanks \cite{dione2019developing}. Details of the treebanks pertinent to parsing can be seen in Table \ref{tab:treebank-stats}. We use pretrained word embeddings from FastText \citep{grave2018learning} for all but Ancient Greek, for which we used embeddings from \citet{ginter2017conll}, and Wolof, for which we used embeddings from \citet{heinzerling2018bpemb}. When necessary, we used the algorithm of \citet{raunak2017simple} to reduce the embeddings to 100 dimensions.

For each treebank we then acquired the following models:
\begin{enumerate}[i]
    \item \textbf{Baseline 1}: Full-sized model is trained as normal and undergoes no compression technique.
    \item \textbf{Baseline 2}: Model is trained as normal but with equivalent sizes of the distilled models (20\%, 40\%, 60\%, and 80\% of the original size) and undergoes no compression technique. These models have the same overall structure of baseline 1, with just the number of dimensions of each layer changed to result in a specific percentage of trainable parameters of the full model.
    \item \textbf{Distilled}: Model is distilled using the \textit{teacher-student} method. We have four models were the first is distilled into a smaller network with 20\% of the parameters of the original, the second 40\%, the third 60\%, and the last 80\%. The network structure and parameters of the distilled models are the exact same as those of the baseline 2 models.
\end{enumerate}
\paragraph{Hardware}
For evaluating the speed of each model when parsing the test sets of each treebank we set the number of CPU cores to be one and either ran the parser using that solitary core or using a GPU (using a single CPU core too). The CPU used was an Intel Core i7-7700 and the GPU was an Nvidia GeForce GTX 1080.\footnote{Using Python 3.7.0, PyTorch 1.0.0, and CUDA 8.0.}

\begin{table*}
    \centering
    \small
    \tabcolsep=.1cm
    \begin{tabular}{l|ccc|ccc|ccc|ccc}
       \multicolumn{1}{c}{}  & \multicolumn{3}{c}{\textbf{number of trees}} & \multicolumn{3}{c}{\textbf{average sent length}}& \multicolumn{3}{c}{\textbf{average arc length}} &
       \multicolumn{3}{c}{\textbf{non-proj. arc pct}} \\
       \multicolumn{1}{c}{}  & \multicolumn{1}{c}{\textbf{train}} & \multicolumn{1}{c}{\textbf{dev}} & \multicolumn{1}{c}{\textbf{test}} & \multicolumn{1}{c}{\textbf{train}} & \multicolumn{1}{c}{\textbf{dev}} & \multicolumn{1}{c}{\textbf{test}} & \multicolumn{1}{c}{\textbf{train}} & \multicolumn{1}{c}{\textbf{dev}} & \multicolumn{1}{c}{\textbf{test}} & \multicolumn{1}{c}{\textbf{train}} & \multicolumn{1}{c}{\textbf{dev}} & \multicolumn{1}{c}{\textbf{test}}  \\\hline\hline&&&&&&&&&&&&\\[-2ex]
   Ancient-Greek-Perseus&11476&1137&1306&14.9&20.5&17.0&4.1&4.5&4.1&23.9&23.2&23.5\\
   Chinese-GSD&3997&500&500&25.7&26.3&25.0&4.7&4.9&4.7&0.1&0.0&0.3\\
   English-EWT&12543&2002&2077&17.3&13.6&13.1&3.7&3.5&3.6&1.0&0.6&0.6\\
   Finnish-TDT&12217&1364&1555&14.3&14.4&14.5&3.4&3.4&3.4&1.6&1.9&1.8\\
   Hebrew-HTB&5241&484&491&27.3&24.6&26.0&3.9&3.8&3.7&0.8&0.8&0.9\\
   Russian-GSD&3850&579&601&20.5&21.2&19.9&3.5&3.7&3.7&1.1&1.0&1.2\\
   Tamil-TTB&400&80&120&16.8&16.8&17.6&3.5&3.7&3.7&0.3&0.0&0.2\\
   Uyghur-UDT&1656&900&900&12.6&12.8&12.5&3.5&3.5&3.5&1.1&1.3&1.4\\
   Wolof-WTB&1188&449&470&20.8&23.9&23.1&3.5&3.8&3.6&0.4&0.4&0.5
    \end{tabular}
    \caption{Statistics for salient features with respect to parsing difficulty for each UD treebank used: number of trees, the number of data instances; average sent length, the length of each data instance on average; average arc length, the mean distance between heads and dependents; non.proj. arc pct, the percentage of non-projective arcs in a treebank.}
    \label{tab:treebank-stats}
\end{table*}

\paragraph{Experiment} We compare the performance of each model on the aforementioned UD treebanks with respect to the unlabelled attachment score (UAS) which evaluates the accuracy of the arcs, and the labelled attachment score (LAS) which also includes the accuracy of the arc labels. We also evaluate the differences in inference time for each model on CPU and GPU with respect to sentences per second and tokens per second. We report sentences per second as this has been the measurement traditionally used \carlos{in most of the literature}, but we also use tokens per second as this more readily captures the difference in speed across parsers for different treebanks where the sentence length varies considerably.  We also report the number of trainable parameters of each distilled model and how they compare to the baseline, as this is considered a good measure of how green a model is in lieu of the number of floating point operations (FPO) \citep{schwartz2019green}.\footnote{There exist a number of packages for computing the FPO of a model but\carlos{, to our knowledge,} as of yet they do not include the capability of dealing with LSTMs.}

\section{Results and Discussion}
\begin{table*}
\centering
\small
\tabcolsep=.067cm
\begin{tabular}{l|cc|cc|cc|cc|cc|cc|cc|cc|cc|cc}
\multicolumn{1}{l}{}&\multicolumn{2}{c}{\textbf{gr}}&\multicolumn{2}{c}{\textbf{zh}}&\multicolumn{2}{c}{\textbf{en}}&\multicolumn{2}{c}{\textbf{fi}}&\multicolumn{2}{c}{\textbf{he}}&\multicolumn{2}{c}{\textbf{ru}}&\multicolumn{2}{c}{\textbf{ta}}&\multicolumn{2}{c}{\textbf{ug}}&\multicolumn{2}{c}{\textbf{wo}}&\multicolumn{2}{c}{\textbf{avg}}\\\multicolumn{1}{l}{}& \multicolumn{1}{c}{UAS} & \multicolumn{1}{c}{LAS}& \multicolumn{1}{c}{UAS} & \multicolumn{1}{c}{LAS}& \multicolumn{1}{c}{UAS} & \multicolumn{1}{c}{LAS}& \multicolumn{1}{c}{UAS} & \multicolumn{1}{c}{LAS}& \multicolumn{1}{c}{UAS} & \multicolumn{1}{c}{LAS}& \multicolumn{1}{c}{UAS} & \multicolumn{1}{c}{LAS}& \multicolumn{1}{c}{UAS} & \multicolumn{1}{c}{LAS}& \multicolumn{1}{c}{UAS} & \multicolumn{1}{c}{LAS}& \multicolumn{1}{c}{UAS} & \multicolumn{1}{c}{LAS}& \multicolumn{1}{c}{UAS} & \multicolumn{1}{c}{LAS}\\\hline\hline& & & & & & & & & & & & & & & & & & & \\[-2ex]
\textbf{Full}&75.5&70.4&88.2&85.9&90.8&89.0&90.5&88.6&90.8&88.6&88.9&85.2&76.9&71.0&75.2&58.9&88.5&84.5&85.0&80.2\\\textbf{B-20}&70.5&64.4&85.1&82.1&88.6&86.4&86.7&83.6&87.9&85.1&86.3&82.0&76.2&69.9&72.2&55.6&86.1&81.8&82.2&76.8\\\textbf{B-40}&72.2&66.4&86.1&83.5&88.9&86.8&87.7&84.8&88.5&85.6&87.1&83.1&78.4&71.8&73.0&55.7&86.5&82.2&83.2&77.8\\\textbf{B-60}&72.0&66.4&86.7&84.0&89.5&87.5&88.1&85.5&88.7&86.3&87.1&83.1&77.5&70.9&72.7&55.9&87.5&83.1&83.3&78.1\\\textbf{B-80}&71.8&66.2&86.7&84.3&89.1&87.1&88.5&85.9&89.3&86.6&87.1&82.9&78.2&71.5&73.0&56.2&87.8&83.6&83.5&78.3\\\hline& & & & & & & & & & & & & & & & & & & \\[-2ex]
\textbf{D-20}&72.3&66.4&86.7&84.2&89.5&87.7&87.6&84.9&89.4&86.7&88.2&84.2&80.6&74.7&74.1&57.9&89.0&85.0&84.1&79.1\\\textbf{D-40}&74.0&68.3&87.9&85.6&89.9&88.0&89.5&86.9&89.4&87.0&88.4&84.6&80.9&74.7&74.5&58.3&89.4&85.5&84.9&79.9\\\textbf{D-60}&74.2&68.7&88.3&85.9&90.1&88.3&89.4&87.1&90.0&87.5&88.6&84.7&80.4&74.5&74.5&58.6&89.5&85.8&85.0&80.1\\\textbf{D-80}&75.0&69.6&88.4&86.2&90.1&88.3&89.2&86.9&90.3&88.0&88.8&85.0&81.2&75.4&74.6&58.6&89.6&85.7&85.3&80.4
\end{tabular}
\caption{Full attachment scores for each model and for each test treebank where Full means the original sized model, B-X means training a model with X\% of the trainable parameters of the original model, and D-X means distilling to a model with X\% of the trainable parameters of the original model.}\label{tab:as-scores}
\end{table*}

\begin{table}[b]
    \centering
    \small
    \tabcolsep=.25cm 
    \begin{tabular}{l|c|c|c|c|c}
\multicolumn{1}{c}{}&    \multicolumn{5}{c}{\textbf{Energy (kJ)}} \\
 \multicolumn{1}{c}{}    & \multicolumn{1}{c}{\textbf{Full}} & \multicolumn{1}{c}{\textbf{D-80}} & \multicolumn{1}{c}{\textbf{D-60}} & \multicolumn{1}{c}{\textbf{D-40}} & \multicolumn{1}{c}{\textbf{D-20}}\\\hline\hline& & & & & \\[-2ex] 
      inf & 0.32 & 0.31 & 0.27 & 0.25 & 0.24 \\
        w/ load & 6.91  & 6.70  & 6.9 &  5.95 &  3.67
    \end{tabular}
    \caption{Total inference energy consumption (inf) used for all test treebanks (8K sentences) and also with the energy consumption used to load each of the 9 models (w/ load). The standard deviation for inference energy consumption was 0.01 exclusively and for the consumption with loading models it ranged from 0.06 to 0.15.}
    \label{tab:inf_energy}
\end{table}
\begin{figure}
\centering
\begin{subfigure}[b]{1.0\linewidth}
\includegraphics[width=\linewidth]{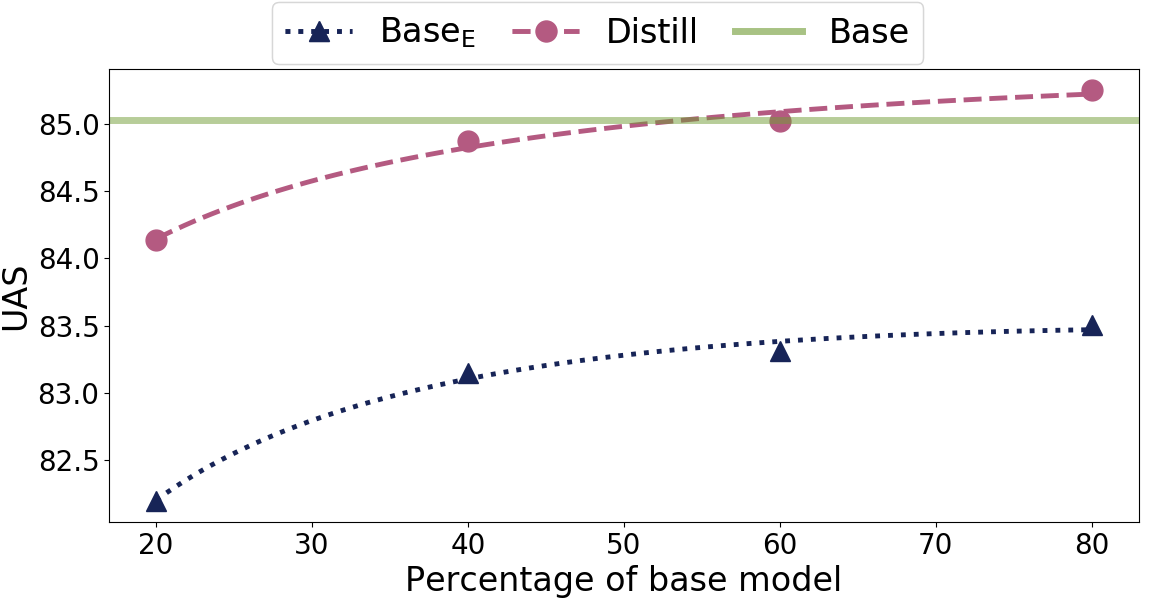}
\caption{}
\label{fig:uas-vs-size}
\end{subfigure}

\begin{subfigure}[b]{1.0\linewidth}
\includegraphics[width=\linewidth]{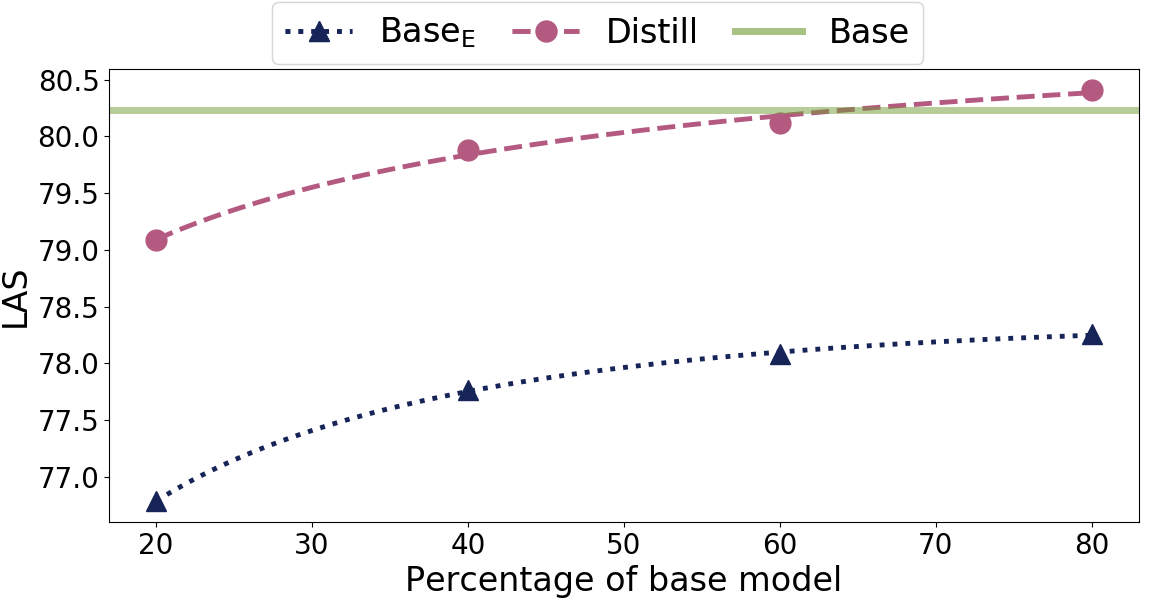}
\caption{}
\label{fig:las-vs-size}
\end{subfigure}
\caption{UAS (a) and LAS (a) against the model size relative to the original full-sized model: Base$_{\textrm{E}}$, the baseline models of equivalent size to the distilled models; Distill, the distilled models; Base, the performance of the original full-sized model.}
\label{fig:as-vs-size}
\end{figure}





\begin{figure}[htbp!]
\centering
\begin{subfigure}{1.0\linewidth}
\includegraphics[width=\linewidth]{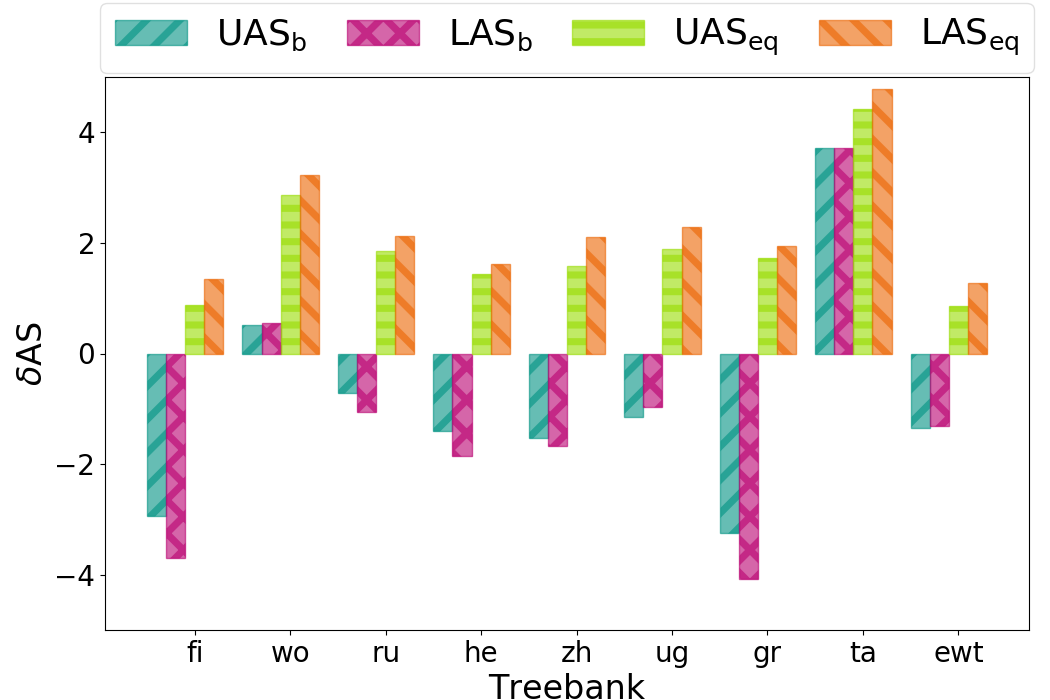}
\caption{}
\label{fig:delta-uas-2}
\end{subfigure}

\begin{subfigure}{1.0\linewidth}
\includegraphics[width=\linewidth]{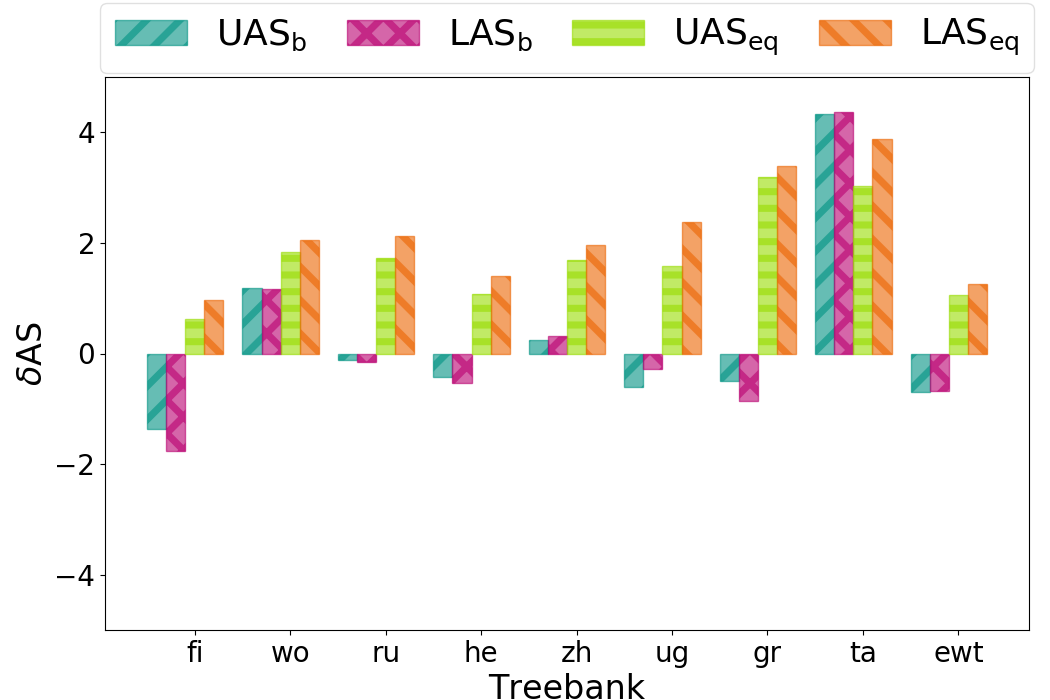}
\caption{}
\label{fig:delta-uas-8}
\end{subfigure}
\caption{Delta UAS and LAS for when comparing both the original base model and equivalent-sized base models for each treebank for two of our distilled models: (a) D-20, 20\% of original model and (b) D-80, 80\% of original model.}
\label{fig:delta-tbs}
\end{figure}

\begin{figure}
\centering
\begin{subfigure}[b]{1.0\linewidth}
\includegraphics[width=\linewidth]{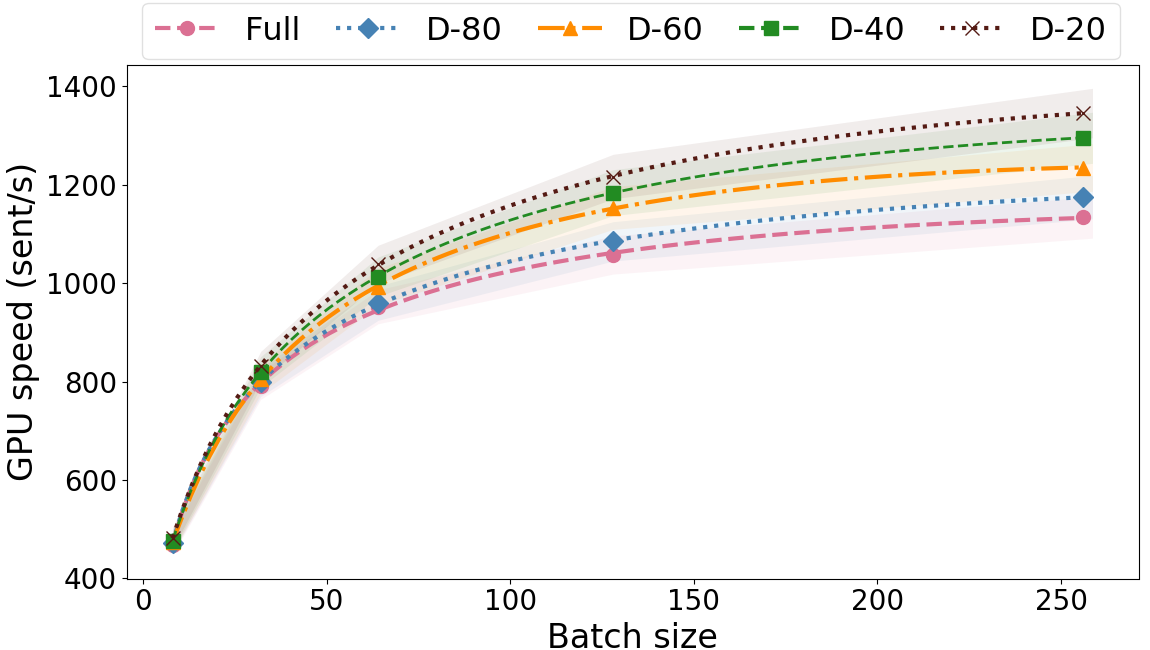}
\caption{}
\label{fig:gpu-speeds}
\end{subfigure}
\begin{subfigure}[b]{1.0\linewidth}
\includegraphics[width=\linewidth]{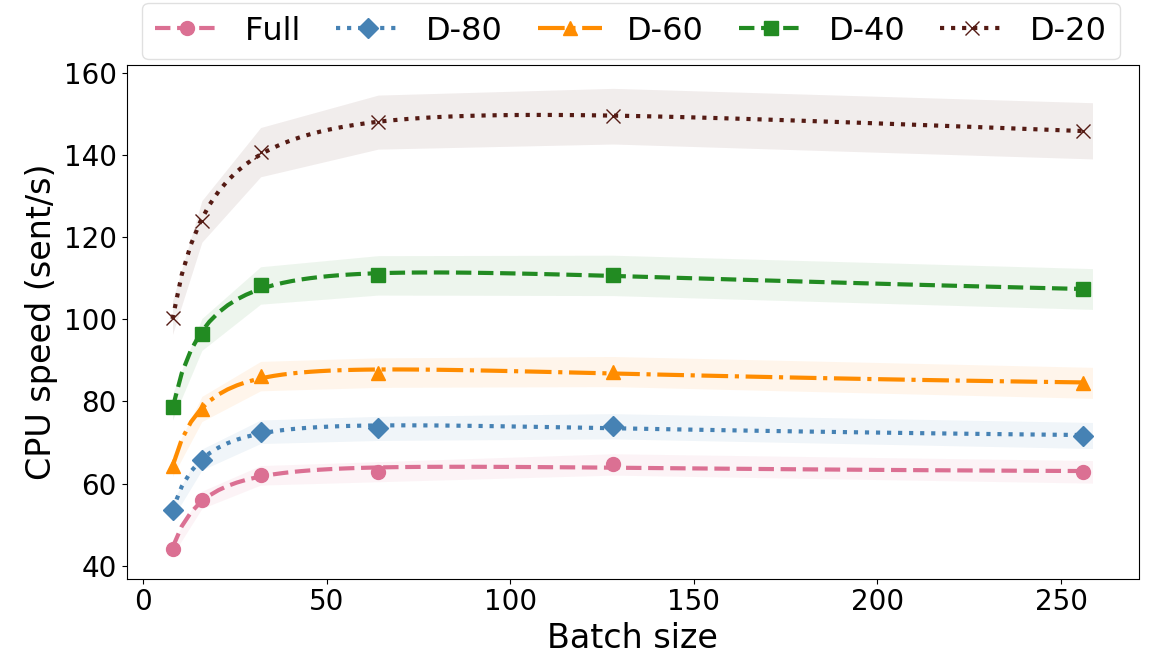}
\caption{}
\label{fig:cpu-speeds}
\end{subfigure}
\caption{GPU (a) and single core CPU (b) speeds in sentence per second with varying batch sizes for distilled models (D-X) and full-sized base model (Full). Shaded areas show the standard error. Speeds for Tamil-TTB are not included as the test treebank is too small for larger batch sizes.}
\label{fig:speeds}
\end{figure}

\begin{figure}
\centering
\begin{subfigure}[b]{1.0\linewidth}
\includegraphics[width=\linewidth]{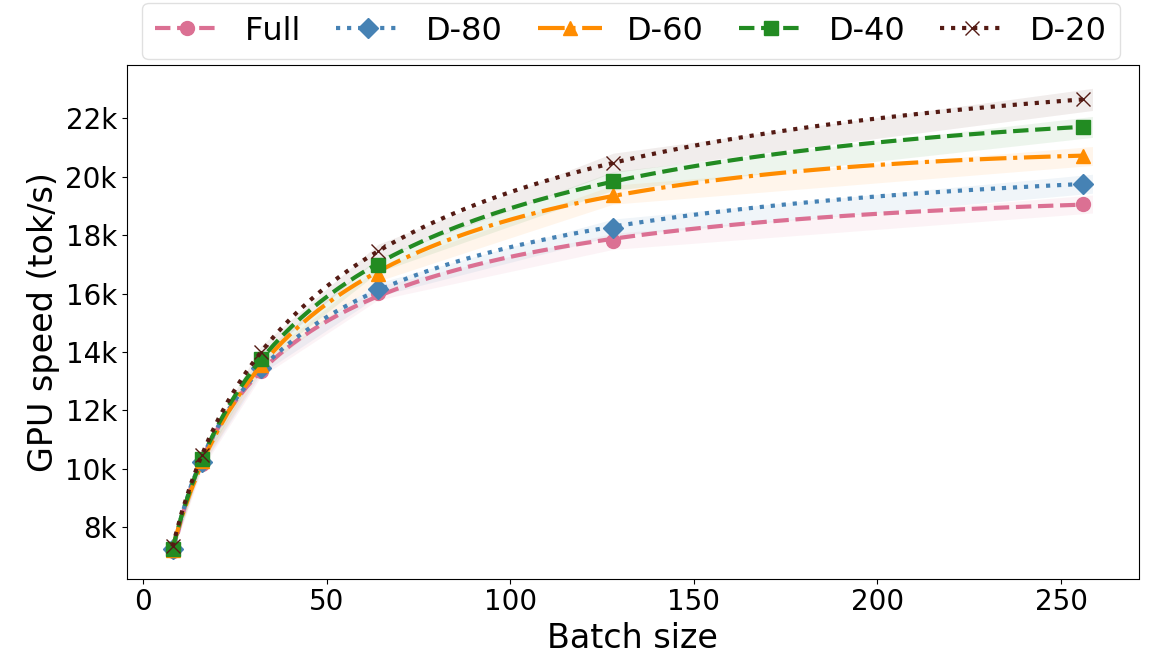}
\caption{}
\label{fig:gpu-speeds-toks}
\end{subfigure}
\begin{subfigure}[b]{1.0\linewidth}
\includegraphics[width=\linewidth]{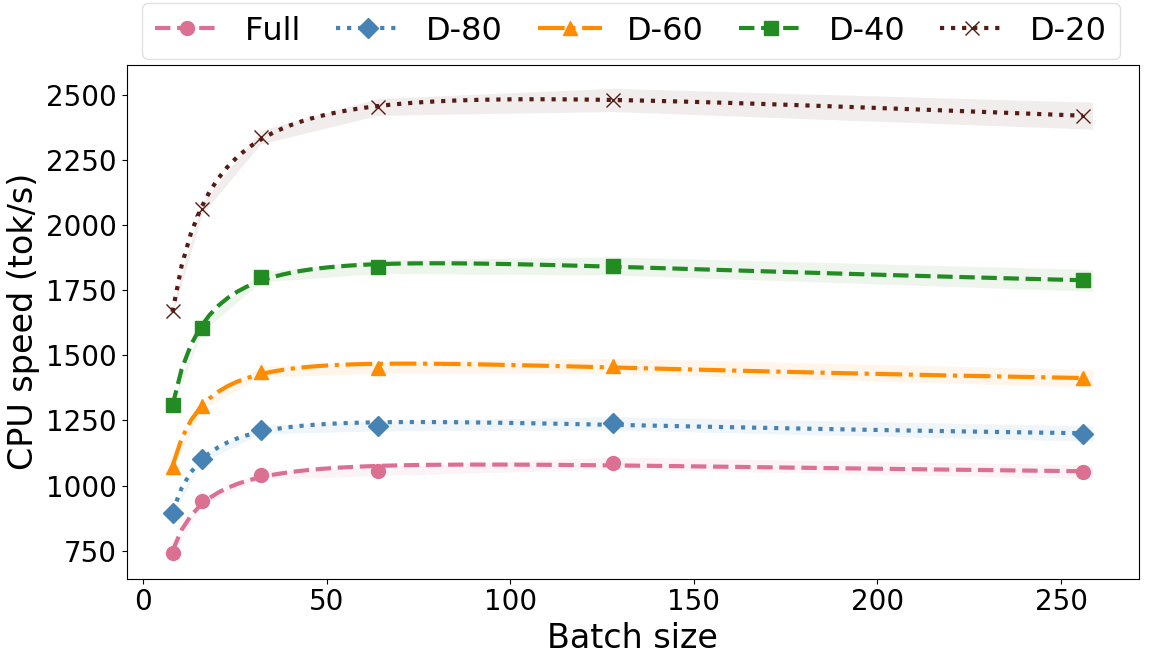}
\caption{}
\label{fig:cpu-speeds-toks}
\end{subfigure}
\caption{GPU (a) and single core CPU (b) speeds in tokens per second with varying batch sizes for distilled models (D-X) and full-sized base model (Full). Shaded areas show the standard error. Speeds for Tamil-TTB are not included as the test treebank is too small for larger batch sizes.}
\label{fig:speeds-toks}
\end{figure}


\begin{figure}
\centering
\begin{subfigure}{1.0\columnwidth}
\includegraphics[width=\columnwidth]{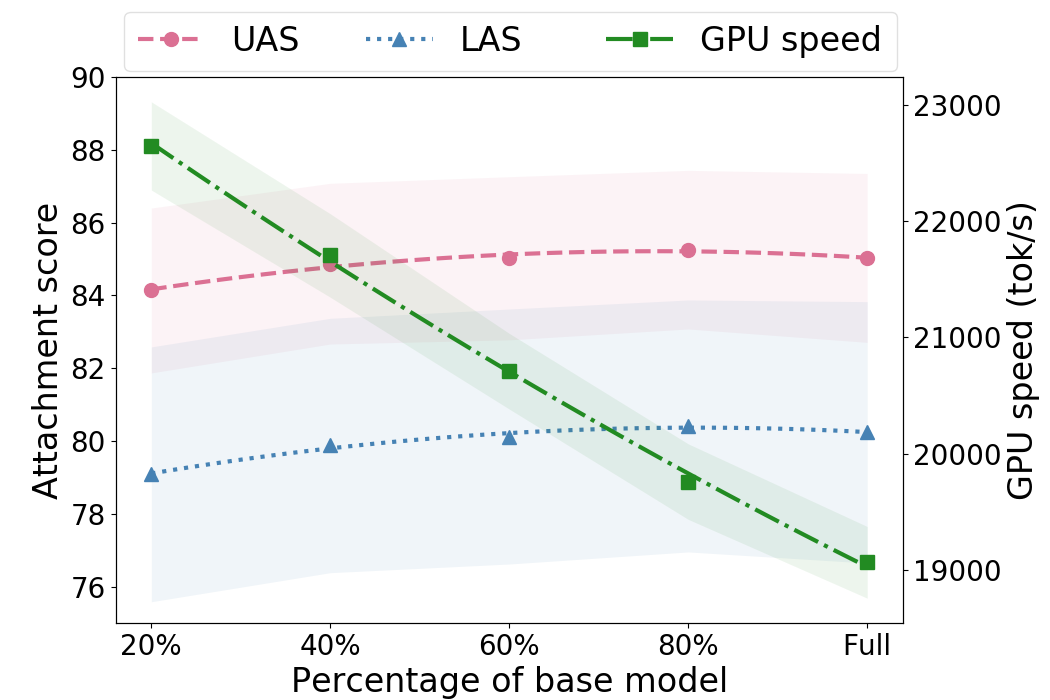}
\caption{}
\label{fig:acc-speed-cpu}
\end{subfigure}
\begin{subfigure}{1.0\columnwidth}
\includegraphics[width=\columnwidth]{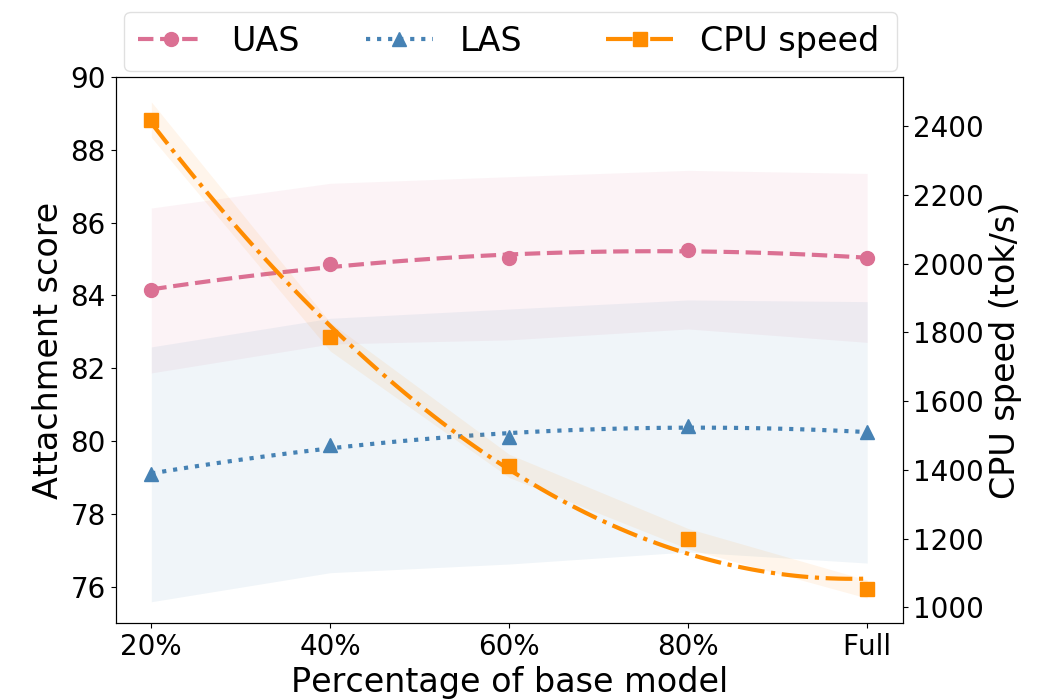}
\caption{}
\label{fig:acc-speed-gpu}
\end{subfigure}

\caption{Comparison of attachment scores and percentage increase of speed (tok/s) for different distilled models with batch size 256: speed on GPU (a) and speed on CPU (b). Shaded areas show the standard error. Speeds for Tamil-TTB are not included as the test treebank is too small for larger batch sizes.}
\label{fig:acc-speed}
\end{figure}
Figure \ref{fig:as-vs-size} shows the average attachment scores across all test treebanks (all results presented in this section are on the test treebanks) for the distilled models and the equivalent-sized base models against the size of the model relative to the original full model. There is a clear gap in performance between these two sets of models with roughly 2 points of UAS and LAS more for the distilled models. This shows that the distilled models do actually manage to leverage the information from the original full model. The full model's scores are also shown and it is clear that on average the model can be distilled to 60\% with no loss in performance. When compressing to 20\% of the full model, the performance only decreases by about 1 point for both UAS and LAS. 

Figures \ref{fig:delta-uas-2} and \ref{fig:delta-uas-8} show the differences in UAS and LAS for the models distilled to 20\% and 80\% respectively for each treebank when compared to the equivalent sized baseline model and the full baseline model. The distilled models far outperform the equivalent-sized baselines for all treebanks.  It is clear that for the smaller model some treebanks suffer more when compressed to 20\% than others when compared to the full baseline model, e.g. Finnish-TDT and Ancient-Greek-Perseus. These two treebanks have the largest percentage of non-projective arcs (as can be seen in Table \ref{tab:treebank-stats}) which could account for the decrease in performance, with a more powerful model required to account for this added syntactic complexity. 

However, the two smallest treebanks, Tamil-TTB and Wolof-WTB, actually increase in accuracy when using distillation, especially Tamil-TTB, which is by far the smallest treebank, with an increase in UAS and LAS of about 4 points over the full base model. This is likely the result of over-fitting when using the larger, more powerful model, so that reducing the model size actually helps with generalisation.

These observations are echoed in the results for the model distilled to 80\%, where most treebanks lose less than a point for UAS and LAS against the full baseline, but have a smaller increase in performance over the equivalent-sized baseline. This makes sense as the model is still close in size to the full baseline and still similarly powerful. The increase in performance for Tamil-TTB and Wolof-WTB are greater for this distilled model, which suggests the full model doesn't need to be compressed to such a small model to help with generalisation. The full set of attachment scores from our experiments can be seen in Table \ref{tab:as-scores}.
\begin{table*}[t]
\centering
\small
\tabcolsep=.075cm  
\begin{tabular}{l|r|r|r|r|r|r|r|r|r}
    \multicolumn{1}{l}{}  & \multicolumn{1}{c}{\textbf{gr}} &
      \multicolumn{1}{c}{\textbf{zh}} &
      \multicolumn{1}{c}{\textbf{en}} &
      \multicolumn{1}{c}{\textbf{fi}} &
      \multicolumn{1}{c}{\textbf{he}} &
      \multicolumn{1}{c}{\textbf{ru}} &
      \multicolumn{1}{c}{\textbf{ta}} &
      \multicolumn{1}{c}{\textbf{ug}} &
      \multicolumn{1}{c}{\textbf{wo}} \\\hline\hline&&&&&&&&&\\[-2.0ex]
\multicolumn{1}{l|}{\textbf{Full}}&\multicolumn{1}{c|}{12.28}&\multicolumn{1}{c|}{11.98}&\multicolumn{1}{c|}{12.23}&\multicolumn{1}{c|}{12.77}&\multicolumn{1}{c|}{12.04}&\multicolumn{1}{c|}{11.92}&\multicolumn{1}{c|}{11.22}&\multicolumn{1}{c|}{11.45}&\multicolumn{1}{c}{11.39}\\\hline&&&&&&&&&\\[-2.0ex]
\textbf{D-20}&2.47 (19.7)&2.42 (20.2)&2.44 (19.7)&2.56 (19.7)&2.39 (19.2)&2.36 (19.3)&2.25 (19.6)&2.30 (20.2)&2.27 (19.5)\\
\textbf{D-40}&4.88 (39.3)&4.79 (39.5)&4.86 (39.3)&5.12 (40.2)&4.80 (40.0)&4.73 (39.5)&4.49 (39.3)&4.60 (40.4)&4.57 (39.8)\\
\textbf{D-60}&7.35 (59.8)&7.24 (60.5)&7.33 (59.8)&7.66 (59.8)&7.19 (59.2)&7.18 (59.7)&6.71 (59.8)&6.90 (60.5)&6.84 (60.2)\\
\textbf{D-80}&9.80 (80.3)&9.57 (79.8)&9.75 (79.5)&10.23 (80.3)&9.59 (79.2)&9.52 (79.8)&8.94 (79.5)&9.19 (79.8)&9.12 (80.5)
\end{tabular}
\caption{Trainable model parameters ($\times10^6$) with percentage of full model in parentheses, where Full means the original sized model and D-X means distilling to a model with X\% of the trainable parameters of the original model.}\label{tab:sizes}
\end{table*}

With respect to how green our distilled models are, Table \ref{tab:sizes} shows the number of trainable parameters for each distilled model for each treebank alongside its corresponding full-scale baseline. We report these in lieu of FPO as\carlos{, to our knowledge,} no packages exist to calculate the FPO for neural network layers like LSTMs which are used in our network. These numbers do not depend on the hardware used and strongly correlate with the amount of memory a model consumes. Different algorithms do utilise parameters differently, however, the models compared here are of the same structure and use the same algorithm, so comparisons of the number of trainable model parameters do relate to how much work each respective model does compared to another. Beyond this we offer a nominal analysis of inference energy consumption for each of the model sizes. These measurements can be seen in Table \ref{tab:inf_energy}. The full baseline uses roughly 33\% more than the smallest distilled model. This difference is more pronounced when including the energy used to load the models (which might be a consideration if the parser cannot be kept in memory) as the full baseline almost uses twice as much energy as the smallest distilled model. 

Figures \ref{fig:speeds} and \ref{fig:speeds-toks} show the parsing speeds on CPU and GPU for the distilled models and for the full baseline model 
in sentences and tokens per second,
respectively. The speeds are reported for different batch sizes as this obviously affects the speed at which a neural network can make predictions, but the maximum batch size that can be used on different systems varies significantly. As can be seen in Figures \ref{fig:gpu-speeds} and \ref{fig:gpu-speeds-toks}, the limiting factor in parsing speed is the bottleneck of loading the data onto the GPU when using a batch size less than $\sim$50 sentences. However, with a batch size of 256 sentences, we achieve an increase in parsing speed of 19\% over the full baseline model when considering tokens per second.

As expected, a much smaller batch size is required to achieve increases in parsing speed when using a CPU. Even with a batch size of 16 sentences, the smallest model more than doubles the speed of the baseline. For a batch size of 256, the distilled model compressed to 20\% increases the speed of the baseline by 130\% when considering tokens per second. A full breakdown of the parsing speeds for each treebank and each model when using a batch size of 256 sentences is given in Table \ref{tab:speed-scores} in the Appendix.

Figure \ref{fig:acc-speed} shows the attachment scores and the corresponding parsing speed against model size for the distilled model and the full baseline model. These plots clearly show that the cost in accuracy is neglible when compared to the large increase in parsing speed. So not only does this \textit{teacher-student} distillation technique maintain the accuracy of the baseline model, but it achieves real compression and with it practical increases in parsing speed and with a greener implementation. In absolute terms, our distilled models are faster than the previously fastest parser using sequence labelling, as can be seen explicitly in Table \ref{tab:current-speeds} for PTB, and outperforms it by over 1 point with respect to UAS and LAS when compressing to 40\%. Distilling to 20\% results in a speed 4x that of the sequence labelling model on CPU but comes at a cost of 0.62 points for UAS and 0.76 for LAS compared to the sequence labelling accuracies. 
Furthermore, the increase in parsing accuracy for the smaller treebanks suggests that distillation could be used as a more efficient way of finding optimal hyperparameters depending on the available data, rather than 
training numerous models with varying hyperparameter settings.

\john{We also need to consider training costs, 
an important factor to implement green AI.
In this respect, while our full baseline model took 66.4 seconds per epoch to train on English-EWT (the largest treebank used in this analysis), the baseline reduced to 20\% trainable parameters required 52.9s per epoch, and the distillation into 
20\% of the original parameters clocked in at 103.1s per epoch.
The distillation process takes longer and must be done after a full model is trained. However, the optimal model when distilling often 
occurred earlier (about epoch 50, rather than 80-100)
suggesting less training is required.}

\john{In practice, the intended use of a parser should be considered when evaluating
the environmental adequacy of distillation:
in systems that will parse at a large scale or be deployed for extended periods of time, the savings at decoding time will offset the increased carbon footprint from training, but this may not be true in smaller-scale scenarios. However, in the latter, distillation can still be useful to reduce hardware requirements of the machine(s) used for decoding, indirectly reducing emissions.}


\subsection{Future work}
There are numerous ways in which this distillation technique could be augmented to potentially retain more performance and even outperform the large baseline models, such as using \textit{teacher annealing} introduced by \citet{clark2019bam} where the distillation process gradually secedes to standard training. 
Beyond this, the structure of the distilled models can be altered, e.g. student models which are more shallow than the teacher models \citep{ba2014}. This technique could further improve the efficiency of models and make them more environmentally friendly by reducing the depth of the models and therefore the total number of trainable parameters. 

Distillation techniques can also be easily expanded to other NLP tasks. Already attempts have been made to make BERT more wieldy by compressing the information it contains into task-specific models \citep{tang2019distilling}. But this can be extended to other tasks more specifically and potentially reduce the environmental impact of NLP research and deployable NLP systems.

\section{Conclusion}
We have obtained results that suggest using \textit{teacher-student} distillation for UD parsing is an effective means of increasing parsing efficiency. The baseline parser used for our experiments was not only accurate but already fast, meaning it was a strong baseline from which to see improvements. We obtained parsing speeds 2.30x (1.19x) faster on CPU (GPU) while only losing $\sim$1 point for both UAS and LAS when compared to the original sized model. Furthermore, the smallest model which obtains these results only has 20\% of the original model's trainable parameters, vastly reducing its environmental impact.
\section*{Acknowledgments}
This work has received funding from the European Research Council (ERC), under the European Union's Horizon 2020 research and innovation programme (FASTPARSE, grant agreement No 714150), from the  ANSWER-ASAP project (TIN2017-85160-C2-1-R) from MINECO, and from Xunta de Galicia (ED431B 2017/01, ED431G 2019/01).
\bibliography{acl2020.bib}
\bibliographystyle{acl_natbib}
\clearpage
\appendix
\onecolumn
\section{Appendix}\label{appendix:sup}
\begin{table}[ht!]
\vspace{-1em}
\newcolumntype{C}{ @{}>{${}}c<{{}$}@{} }
\centering
\footnotesize
\tabcolsep=.1cm
\begin{tabular}{p{0.5em}p{1.2em}c|*4{rCl|}*1{rCl}}
      \multicolumn{1}{c}{} & \multicolumn{1}{c}{} & \multicolumn{1}{c}{}  & \multicolumn{3}{c}{\textbf{Full}} & \multicolumn{3}{c}{\textbf{D-20}} & \multicolumn{3}{c}{\textbf{D-40}} & \multicolumn{3}{c}{\textbf{D-60}} & \multicolumn{3}{c}{\textbf{D-80}} \\[0.5ex] \hline\hline & & & & & & & & & & & & & & &\\[-2ex]
\multirow{4}{*}{\textbf{gr}} & \multirow{2}{*}{CPU} & (tok/s)&1211& \pm &2&2842& \pm &3&2086& \pm &3&1638& \pm &3&1390& \pm &1\\
 && (sent/s)&75.4& \pm &0.1&177.1& \pm &0.2&130.0& \pm &0.2&102.1& \pm &0.2&86.6& \pm &0.0\\
 & \multirow{2}{*}{GPU} & (tok/s)&19219& \pm &77&21017& \pm &142&21296& \pm &122&20346& \pm &70&19202& \pm &147\\
 && (sent/s)&1197.6& \pm &4.8&1309.6& \pm &8.9&1327.0& \pm &7.6&1267.8& \pm &4.3&1196.5& \pm &9.1\\\hline
\multirow{4}{*}{\textbf{zh}} & \multirow{2}{*}{CPU} & (tok/s)&1124& \pm &2&2503& \pm &3&1872& \pm &2&1490& \pm &2&1278& \pm &1\\
 && (sent/s)&46.8& \pm &0.1&104.2& \pm &0.1&77.9& \pm &0.1&62.0& \pm &0.1&53.2& \pm &0.0\\
 & \multirow{2}{*}{GPU} & (tok/s)&21255& \pm &113&25665& \pm &82&24862& \pm &134&23567& \pm &28&22663& \pm &91\\
 && (sent/s)&884.7& \pm &4.7&1068.3& \pm &3.4&1034.9& \pm &5.6&981.0& \pm &1.2&943.4& \pm &3.8\\\hline
\multirow{4}{*}{\textbf{en}} & \multirow{2}{*}{CPU} & (tok/s)&884& \pm &1&2217& \pm &10&1548& \pm &3&1217& \pm &1&1010& \pm &7\\
 && (sent/s)&73.2& \pm &0.1&183.5& \pm &0.8&128.1& \pm &0.3&100.7& \pm &0.1&83.6& \pm &0.6\\
 & \multirow{2}{*}{GPU} & (tok/s)&16942& \pm &25&20538& \pm &60&19739& \pm &109&19003& \pm &90&17511& \pm &57\\
 && (sent/s)&1402.2& \pm &2.1&1699.8& \pm &5.0&1633.7& \pm &9.0&1572.7& \pm &7.4&1449.2& \pm &4.7\\\hline
\multirow{4}{*}{\textbf{fi}} & \multirow{2}{*}{CPU} & (tok/s)&988& \pm &1&2586& \pm &3&1767& \pm &2&1371& \pm &2&1153& \pm &0\\
 && (sent/s)&72.9& \pm &0.0&190.9& \pm &0.2&130.4& \pm &0.2&101.2& \pm &0.1&85.1& \pm &0.0\\
 & \multirow{2}{*}{GPU} & (tok/s)&18325& \pm &46&22181& \pm &50&21408& \pm &130&20220& \pm &90&19013& \pm &33\\
 && (sent/s)&1352.4& \pm &3.4&1637.0& \pm &3.7&1580.0& \pm &9.6&1492.3& \pm &6.7&1403.2& \pm &2.4\\\hline
\multirow{4}{*}{\textbf{he}} & \multirow{2}{*}{CPU} & (tok/s)&1180& \pm &1&2644& \pm &3&1964& \pm &2&1582& \pm &1&1337& \pm &1\\
 && (sent/s)&47.2& \pm &0.0&105.7& \pm &0.1&78.5& \pm &0.1&63.3& \pm &0.0&53.5& \pm &0.0\\
 & \multirow{2}{*}{GPU} & (tok/s)&22202& \pm &98&26441& \pm &150&25418& \pm &181&24233& \pm &176&22651& \pm &89\\
 && (sent/s)&887.4& \pm &3.9&1056.8& \pm &6.0&1016.0& \pm &7.2&968.6& \pm &7.1&905.4& \pm &3.5\\\hline
\multirow{4}{*}{\textbf{ru}} & \multirow{2}{*}{CPU} & (tok/s)&734& \pm &1&1717& \pm &3&1237& \pm &1&976& \pm &1&832& \pm &1\\
 && (sent/s)&38.7& \pm &0.0&90.6& \pm &0.1&65.3& \pm &0.1&51.5& \pm &0.1&43.9& \pm &0.1\\
 & \multirow{2}{*}{GPU} & (tok/s)&16383& \pm &87&19661& \pm &137&18337& \pm &44&17901& \pm &65&17014& \pm &21\\
 && (sent/s)&864.9& \pm &4.6&1037.9& \pm &7.2&968.0& \pm &2.3&944.9& \pm &3.4&898.2& \pm &1.1\\\hline
\multirow{4}{*}{\textbf{ta}} & \multirow{2}{*}{CPU} & (tok/s)&1110& \pm &2&2334& \pm &5&1799& \pm &1&1464& \pm &2&1251& \pm &2\\
 && (sent/s)&67.0& \pm &0.1&140.8& \pm &0.3&108.5& \pm &0.1&88.3& \pm &0.1&75.5& \pm &0.1\\
 & \multirow{2}{*}{GPU} & (tok/s)&17188& \pm &194&19829& \pm &126&19771& \pm &106&18540& \pm &98&18172& \pm &151\\
 && (sent/s)&1037.0& \pm &11.7&1196.3& \pm &7.6&1192.8& \pm &6.4&1118.6& \pm &5.9&1096.4& \pm &9.1\\\hline
\multirow{4}{*}{\textbf{ug}} & \multirow{2}{*}{CPU} & (tok/s)&1058& \pm &1&2289& \pm &3&1806& \pm &2&1404& \pm &2&1199& \pm &2\\
 && (sent/s)&92.2& \pm &0.1&199.4& \pm &0.3&157.3& \pm &0.2&122.4& \pm &0.2&104.5& \pm &0.1\\
 & \multirow{2}{*}{GPU} & (tok/s)&17974& \pm &35&21298& \pm &82&21004& \pm &93&19738& \pm &70&18963& \pm &132\\
 && (sent/s)&1566.0& \pm &3.0&1855.6& \pm &7.2&1829.9& \pm &8.1&1719.6& \pm &6.1&1652.1& \pm &11.5\\\hline
\multirow{4}{*}{\textbf{wo}} & \multirow{2}{*}{CPU} & (tok/s)&1245& \pm &2&2559& \pm &5&2021& \pm &3&1614& \pm &2&1398& \pm &2\\
 && (sent/s)&56.3& \pm &0.1&115.6& \pm &0.2&91.3& \pm &0.1&72.9& \pm &0.1&63.2& \pm &0.1\\
 & \multirow{2}{*}{GPU} & (tok/s)&20225& \pm &74&24361& \pm &94&21564& \pm &73&20661& \pm &102&21059& \pm &105\\
 && (sent/s)&913.8& \pm &3.4&1100.6& \pm &4.2&974.2& \pm &3.3&933.4& \pm &4.6&951.4& \pm &4.7\\\hline
\multirow{4}{*}{\textbf{avg}} & \multirow{2}{*}{CPU} & (tok/s)&1070& \pm &21&2440& \pm &39&1808& \pm &32&1431& \pm &26&1218& \pm &23\\
 && (sent/s)&63.5& \pm &2.1&146.8& \pm &5.4&108.1& \pm &3.8&85.3& \pm &2.9&72.5& \pm &2.4\\
 & \multirow{2}{*}{GPU} & (tok/s)&18933& \pm &243&22503& \pm &307&21488& \pm &271&20463& \pm &251&19666& \pm &252\\
 && (sent/s)&1124.7& \pm &33.3&1336.3& \pm &40.1&1282.8& \pm &41.7&1220.4& \pm &38.8&1168.2& \pm &34.8\\\hline

\end{tabular}
\caption{Speeds with batch size 256.}\label{tab:speed-scores}
\end{table}

\begin{table}[H]
\footnotesize
    \centering
        \tabcolsep=.25cm  
    \begin{tabular}{l p{3.5em} r}
    \textbf{hyperparameter} & & \textbf{value}\\
    \hline\hline& &\\[-2ex]
         word embedding dimensions& & 100\\
         pos embedding dimensions&  & 100\\
         embedding dropout&  & 0.33 \\
         BiLSTM dimensions&  & 400  \\
         BiLSTM layers&  & 3 \\
         arc MLP dimensions&  & 500\\
         label MLP dimensions&  & 100\\
         MLP layers&  & 1 \\
         learning rate&  & 0.2 \\
         dropout&  & 0.33 \\
         momentum&  & 0.9 \\
         L2 norm $\lambda$&  & 0.9\\
         annealing&  & 0.75$^{\wedge}(\nicefrac{t}{5000})$\\
         $\epsilon$& & 1$\times 10^{-12}$\\
         optimiser&  & Adam \\
         loss function&  & cross entropy \\
         epochs&  & 100
    \end{tabular}
    \caption{Hyperparameters for full-sized baseline models.}
    \label{tab:experimental_hyperparameters}
\end{table}

\end{document}